\documentclass{article}

\usepackage{arxiv}

\usepackage[utf8]{inputenc} % allow utf-8 input
\usepackage[T1]{fontenc}    % use 8-bit T1 fonts
\usepackage{enumitem}
\usepackage{hyperref}       % hyperlinks
\usepackage{url}            % simple URL typesetting
\usepackage{booktabs}       % professional-quality tables
\usepackage{amsfonts}       % blackboard math symbols
\usepackage{nicefrac}       % compact symbols for 1/2, etc.
\usepackage{microtype}      % microtypography
\usepackage{lipsum}
\usepackage{graphicx}
\usepackage{amsmath}
\usepackage{verbatim}
\usepackage{framed}
\usepackage{lettrine}
\usepackage[numbers]{natbib}
\graphicspath{ {./images/} }

\title{Exploring Content Based Image Retrieval For Highly Imbalanced Melanoma Data Using Style Transfer, Semantic Image Segmentation and Ensemble Learning}
\author{
 Priyam Mehta\\
    Dep. of Information Technology,\\
    Gujarat Technological University\\
    Gujarat, India \\
    \texttt{priyammehta145@gmail.com}
}

\begin{document}
\maketitle
\begin{abstract}
Lesion images are frequently taken in open-set settings. Because of this, the image data generated is extremely varied in nature.It is difficult for a convolutional neural network to find proper features and generalise well, as a result content based image retrieval (CBIR) system for lesion images are difficult to build. This paper explores this domain and proposes multiple similarity measures which uses Style Loss and Dice Coefficient via a novel similarity measure called I1-Score. Out of the CBIR similarity measures proposed, pure style loss approach achieves a remarkable accuracy increase over traditional approaches like Euclidean Distance and Cosine Similarity. The I1-Scores using style loss performed better than traditional approaches by a small margin, whereas, I1-Scores with dice-coefficient faired very poorly. The model used is trained using ensemble learning for better generalization.  
\end{abstract}

% keywords can be removed

\section{Introduction}
\lettrine{E}{xplosion} in popularity of Deep Learning models has led to more and more services making use of this versatile technology to reduce the burden on human workforce. Deep Learning has penetrated all the sectors, but the momentum of adoption seen in medical domain is very less relative to the other sectors. This slow pace of adoption has been linked to humans' lack of trust in AI models (\citet{trustinhumans}). This mistrust will increase as more people connect to internet and use publicly available models for medical purposes which do not yield good results. One of the most popular example is patients using Google Reverse Image Search to perform differential diagnosis by uploading an image of their lesion on Google Search. Research was conducted by \citet{10.1001/jamadermatol.2016.2096} where it was found that Google's visual object recognition algorithm performed well with objects like chair, cat, dog but it didn't fair well with lesion images because biological images require domain knowledge which was difficult for a general visual object recognition algorithm to perceive . Hence, there is a need for domain specific Content Based Image Retrieval(CBIR) models when dealing with biological images. There have been advances in medical domain specific CBIR models like Similar Image Search for Histopathy(SMILY) by \citet{Hegde_2019}, Deep learning based image retrieval model by \citet{Otalora408237} and \citet{Qayyum_2017} etc. This paper explores CBIR on highly imbalanced melanoma data using style transfer based and semantic segmentation based similarity measures with the help of a novel similarity measure called I1-Score and compares the measures based on the 2 concepts with the existing approaches. I1-Score can incorporate multiple similarity measures under one-roof.

Style Transfer is a neural network based approach developed by \citet{Gatys_2016_CVPR} for transferring texture from one image to another image. It is capable of capturing the texture pecularities like color combinations, brush strokes' size and length etc. Hence, for this research, it was hypothesized that using style transfer in CBIR can help in capturing color, texture, spread and background similarities of the lesions with a lot more accuracy than traditional approaches like Euclidean Distance and Cosine Similarity. In this research, a neural network based classification ensemble was used to get as much generalized learning as possible.

Semantic Image Segmentation is a technique of labeling each pixel of an image with a corresponding class. Research has already been conducted for Image Segmentation based CBIR systems like enhancing the color features by implicitly incorporating shape information from prior segmentation by \citet{bose2015hybrid} or narrowing the search space of similar images using deep learning model by \citet{doi:10.1063/5.0004087}, to increase retrieval accuracy of CBIR systems. For the research here, it was hypothesized that for majority of the melanoma images, the lesion boundaries are blurry, so using semantic image segmentation will enhance features like shape, size and location(relative to image) and thus aid in retrieving more accurate similar images. For the approach here, deep learning based Semantic Segmentation model was used because of their ability to handle complex scenarios as melanoma data was extremely variable in nature due to no guidelines being present for taking pictures of lesions or wounds. 

This paper is organized as follows. Section 2 explains the novel similarity measure called I1-Score. Section 3 presents the research pipeline used for this experiment. Section 4 details the evaluation strategy and shows the results and Section 5 summarizes the effectiveness of novel similarity measure.

\section{Similarity Measures}
The research was conducted using 2 traditional approachs, 1 novel approach, and 6 different similarity measure combination which used I1-Score approach. They are Euclidean Distance, Cosine Similarity, Style Loss, I1\textsubscript{ED} (Euclidean-Dice), I1\textsubscript{CD} (Cosine-Dice), I1\textsubscript{ECD} (Euclidean-Cosine-Dice), I1\textsubscript{SD} (Style-Dice), I1\textsubscript{SE} (Style-Euclidean) and I1\textsubscript{SC} (Style-Cosine). $I1$ score is extremely similar to $F1$ score. Since, this metric is going to be used for Reverse Image Search (RIS), it is named $I1$, where ‘$I$’ refers to ‘Image’.

Style Transfer contains 3 losses - Style Loss, Content Loss, Total Variation Loss. Style Loss helps in determining the quality of texture transfer, Content Loss helps in determining the quality of content retention, and Total Variation Loss helps in determining the overall loss (addition of Style and Content Losses). Only Style Loss was chosen for this research as the main focus is on increasing the importance of texture features. Two lesions can be counted as similar if their color and texture are similar even though the shape and boundaries are extremely different, but the opposite is not true and style loss can help us with that.

Semantic Segmentation of an image relies heavily on features like  color, texture, size, shape and location (relative to image) for proper segregation of pixels but when comparing two semanticaly segmented images only size, shape and location(relative to image) are taken into account as the color and texture would not be present in a semantically segmented image. Hence, using Dice Coefficient will increase the importance of size, shape and location(relative to image).

$I1$ Score combines Dice Coefficient and Style Loss with Euclidean Distance or Cosine Similarity using the same mathematical intuition as the $F1$ Score i.e. only when all the similarity measures used are high, will the overall similarity measure be high.

\subsection{Euclidean Distance}
\begin{gather}
    %\label{}
    ED(x, y) = \sqrt{\sum_{i=0}^{n}(x_{i} - y_{i})^2}\\
    \begin{aligned}
        \text{where,}\\
        & ES(x, y) \text{ = Euclidean Score,}\\
        & ED(x, y) \text{ = Euclidean Distance,}\\
        & ED_{max} \text{ = Max Euclidean Distance,}\\
        & n \text{ = feature vector length,}\\
        & x, y \text{ = Feature vectors from classification model}\\
    \end{aligned}
    \notag
\end{gather}

Euclidean distance is a measure of distance having a unit and follows the principle - “higher the value, less the similarity” whereas to use any metric in I1 Score, it should follow the rule “higher the value, more the similarity”, hence, it is not directly usable.

Converting euclidean distance to euclidean score:
    \begin{enumerate}
        \item Find the euclidean distance of all the images in the database from the input image.
        \item Find the maximum euclidean distance ($ED_{max}$). 
        \item Divide all the distances by $ED_{max}$ to get $ED_{ratio}$. This makes the range [0,1].
        \item Subtract $ED_{ratio}$ from 1, i.e. perform $1 - ED_{ratio}$ to get Euclidean Score or $ES$.
    \end{enumerate}
    We can combine steps 3 and 4 using the below given equation:
    
    %%\begin{equation}
        
    %%\end{equation}
    \begin{gather}
        %\label{eq:1.1}
        ES(x, y) = 1 - \frac{ED(x, y)}{ED_{max}}\\
        \begin{aligned}
            \text{where,}\\
            & ES(x, y) \text{ = Euclidean Score,}\\
            & ED(x, y) \text{ = Euclidean Distance,}\\
            & ED_{max} \text{ = Max Euclidean Distance,}\\
            & x, y \text{ = Feature vectors from classification model}\\
        \end{aligned}
        \notag
    \end{gather}

\subsection{Cosine Similarity}
\begin{gather}
    CS(x, y) = \frac{x.y}{||x||.||y||}\\
    \begin{aligned}
        \text{where,}\\
        & ES(x, y) \text{ = Euclidean Score,}\\
        & ED(x, y) \text{ = Euclidean Distance,}\\
        & ED_{max} \text{ = Max Euclidean Distance,}\\
        & x, y \text{ = Feature vectors from classification model}\\
    \end{aligned}
    \notag
\end{gather}
 Also, Cosine Similarity is a function whose range is [-1, 1], so, it is also not directly usable. To make the measures usable, below given steps were performed.
 
 Converting the range of cosine similarity: 
    \begin{enumerate}
        \item Find cosine similarity between input image (x) and an image in database (y).
        \item Convert range using below given function: 
            \begin{gather}
                %\label{eq:1.3}
                CS_{new}(x, y) = \frac{CS(x, y) + 1}{2}\\
                \begin{aligned}
                    \text{where,}\\
                    & CS_{new}(x, y) \text{ = New Cosine Similarity,}\\
                    & CS(x, y) \text{ = Original Cosine Similarity,}\\
                    & x, y \text{ = Feature vectors from classification model}\\
                \end{aligned}
                \notag
            \end{gather}
            The above formula is derived using the below given formula for converting range of any function.
            \begin{gather}
                V_{new} = \frac{(V_{old} - RO_{min}) * (RN_{max} - RN_{min})}{(RO_{max} - RO_{min}) * RN_{min}}\\
                \begin{aligned}
                    \text{where,}\\
                    & V_{new} \text{ = New Value after range conversion, }\\
                    & V_{old} \text{ = Original or Old value,} \\
                    & RO_{min} \text{ = Minimum value of original function,}\\
                    & RO_{max} \text{ = Maximum value of original function,}\\
                    & RN_{min} \text{ = New lower limit,}\\
                    & RN_{max} \text{ = New upper limit,}\\
                \end{aligned}
                \notag
            \end{gather}
    \end{enumerate}
 
\subsection{Style Loss}
In Style Loss, we pull the feature maps from the chosen hidden layers and create a Gram Matrix for each layer whose feature maps we have pulled. These Gram Matrices get compared. Gram Matrix is a way of representing all the feature maps of a particular layer in a 2D Matrix without losing information or we can say that a set of feature maps gets compressed from 3D to 2D without excessive information loss. The 3D Feature map matrix is flattened along the 1\textsuperscript{st} and 2\textsuperscript{nd} axis and then multiplied with its transpose to get a gram matrix. In style transfer, The gram matrices of generated image and style image are used for calculating the style loss but here, the generated image will be replaced with input image and the style image will be replaced with the image under question (similar or not).

\begin{gather}
    %\label{}
    LS(G^{x}, G^{y}) = \sum_{l}\sum_{i,j}( \beta G_{i,j}^{x,l} - \beta G_{i,j}^{y,l} )^2\\
    \begin{aligned}
        \text{where,}\\
        & LS(G^{x}, G^{y}) \text{ = Style Loss,}\\
        & G^{x}, G^{y} \text{ = Set of Gram Matrices of Input image and Image in question respectively,}\\ 
        & l \text{ = Layer whose gram matrix are currently being processed,}\\
        & \beta \text{ = Weight of the gram matrix of layer $l$,}\\
    \end{aligned}
    \notag
\end{gather}

Similar to Euclidean distance, Style loss is also not bounded. Hence, it needs to be converted to have a range [0,1] to be used in I1-Score. The steps are similar to Euclidean Distance.

Converting style loss to style score:
    \begin{enumerate}
        \item Find the style loss of all the images in the database from the input image.
        \item Find the maximum style loss ($LS_{max}$). 
        \item Divide all the style losses by $LS_{max}$ to get $LS_{ratio}$. This makes the range [0,1].
        \item Subtract $LS_{ratio}$ from 1, i.e. perform $1 - LS_{ratio}$ to get Style Score or $SS$.
    \end{enumerate}
    We can combine steps 3 and 4 using the below given equation:
    
    %%\begin{equation}
        
    %%\end{equation}
    \begin{gather}
        %\label{eq:1.1}
        SS(G^{x}, G^{y}) = 1 - \frac{LS(G^{x}, G^{y})}{LS_{max}}\\
        \begin{aligned}
            \text{where,}\\
            & SS(G^{x}, G^{y}) \text{ = Style Score,}\\
            & G^{x}, G^{y} \text{ = Set of Gram Matrices of Input image and Image in question respectively,}\\
            & LS(G^{x}, G^{y}) \text{ = Style loss between Input image and Image in question}\\
            & LS_{max} \text{ = Maximum Style loss}
        \end{aligned}
        \notag
    \end{gather}

Sørensen-Dice Coefficient is a ratio of same units and also has a range [0,1]. Hence, no changes needed. 

\subsection{Sørensen-Dice Coefficient}
\begin{gather}
    %\label{eq:1.3}
    DC(X, Y) = \frac{2| X \cap Y |}{|X| + |Y|}\\
    \begin{aligned}
        \text{where,}\\
        & DC(X, Y) \text{ = Sørensen-Dice Coefficient,}\\
        & X, Y \text{ = Image output of segmentation model}\\
    \end{aligned}
    \notag
\end{gather}

Sørensen-Dice Coefficient is a ratio of same units and also has a range [0,1]. Hence, no changes needed.

\subsection{Combining similarity measures}
All three I1-Scores use Euclidean Score ($ES$) and New Cosine Similarity ($CS_{new}$).
For below given equations:\\
$x_{c}$, $y_{c}$ = Features vectors of input image and an image from database\\
$x_{s}$ , $y_{s}$ = Segmented images of input image and an image from database\\
$G^{x}$, $G^{y}$ = Gram Matrices of input image and an image from database\\ 
$ES(x, y)$ = Euclidean Score\\
$CS_{new}(x, y)$ = Cosine Similarity with range [0,1]\\
$SS(G^{x}, G^{y})$ = Style Score\\
$DC(X, Y)$ = Sørensen-Dice Coefficient

\subsubsection{I1\texorpdfstring{\textsubscript{ED}}{} Score}
\begin{gather}
    I1_{ED}(x_{c}, y_{c}, x_{s}, y_{s}) = \frac{2}{(\frac{1}{ES(x_{c}, y_{c})}+\frac{1}{DC(x_{s}, y_{s})})}\\
    \begin{aligned}
        \text{where,}\\
        & I1_{ED} \text{ = I1 Score using Euclidean Score and Dice Coefficient}\\
    \end{aligned}
    \notag
\end{gather}

\subsubsection{I1\texorpdfstring{\textsubscript{CD}}{} Score}
\begin{gather}
    I1_{CD}(x_{c}, y_{c}, x_{s}, y_{s}) = \frac{2}{(\frac{1}{CS_{new}(x_{c}, y_{c})}+\frac{1}{DC(x_{s}, y_{s})})}\\
    \begin{aligned}
        \text{where,}\\
        & I1_{CD} \text{ = I1 Score using Cosine Similarity and Dice Coefficient}\\
    \end{aligned}
    \notag
\end{gather}

\subsubsection{I1\texorpdfstring{\textsubscript{ECD}}{} Score}
\begin{gather}
    I1_{ECD}(x_{c}, y_{c}, x_{s}, y_{s}) = \frac{3}{(\frac{1}{ES(x_{c}, y_{c})}+\frac{1}{CS_{new}(x_{c}, y_{c})}+\frac{1}{DC(x_{s}, y_{s})})}\\
    \begin{aligned}
        \text{where,}\\
        & I1_{ECD} \text{ = I1 Score using Euclidean Score, Cosine Similarity and Dice Coefficient}\\
    \end{aligned}
    \notag
\end{gather}

\subsubsection{I1\texorpdfstring{\textsubscript{SE}}{} Score}
\begin{gather}
    I1_{SE}(x_{c}, y_{c}, G^{x}, G^{y}) = \frac{2}{(\frac{1}{ES(x_{c}, y_{c})}+\frac{1}{SS(G^{x}, G^{y})})}\\
    \begin{aligned}
        \text{where,}\\
        & I1_{SE} \text{ = I1 Score using Style Score and Euclidean Distance}\\
    \end{aligned}
    \notag
\end{gather}

\subsubsection{I1\texorpdfstring{\textsubscript{SC}}{} Score}
\begin{gather}
    I1_{SC}(x_{c}, y_{c}, G^{x}, G^{y}) = \frac{2}{(\frac{1}{SS(G^{x}, G^{y})}+\frac{1}{CS_{new}(x_{c}, y_{c})})}\\
    \begin{aligned}
        \text{where,}\\
        & I1_{SC} \text{ = I1 Score using Style Score and Cosine Similarity}\\
    \end{aligned}
    \notag
\end{gather}

\subsubsection{I1\texorpdfstring{\textsubscript{SD}}{} Score}
\begin{gather}
    I1_{SD}(G^{x}, G^{y}, x_{s}, y_{s}) = \frac{2}{(\frac{1}{SS(G^{x}, G^{y})}+\frac{1}{DC(x_{s}, y_{s})})}\\
    \begin{aligned}
        \text{where,}\\
        & I1_{SD} \text{ = I1 Score using Style Score and Dice Coefficient}\\
    \end{aligned}
    \notag
\end{gather}

\section{Research Pipeline}
This section discusses the entire pipeline of research in a step by step manner. Research pipeline consists of 5 main parts: 1) Data Acquisition, 2) Data Pre-processing, 2) Classification Model, 3) Segmentation Model, 4) RIS Module.

\subsection{Data Acquisition}
For research, two datasets were built - Classification dataset and Segmentation dataset. Classification dataset was built by combining \textbf{ISIC}-\textbf{2017} \cite{codella2018skin}, \textbf{2018} \cite{codella2019skin},\cite{Tschandl_2018}, \textbf{2019} \cite{Tschandl_2018},\cite{codella2018skin},\cite{combalia2019bcn20000}, \textbf{2020} \cite{rotemberg2020patientcentric} datasets and Segmentation dataset was built by combining \textbf{PH\textsuperscript{2}}\cite{6610779} and \textbf{ISIC-2017} \cite{codella2018skin} (Segmentation set) datasets.

\subsection{Data Pre-processing}
Images were of varying sizes, hence all the images were resized to have dimension 384x384. Both the datasets also underwent a robust augmentation pipeline. The augmentation pipeline consisted of 7 augmentation methods. They were:
\begin{itemize}
    \item Grid Mask Augmentation by \citet{chen2020gridmask}
    \item Random Left or Right flip
    \item Random Up or Down flip
    \item Random Brightness
    \item Random Contrast
    \item Random Hue
    \item Random Saturation
\end{itemize}
So, for each image there would be 7 more different images, thus increasing the size of the datasets by 7 folds. This also introduced large variations in the datasets, which resulted in better performance on test set. Both the datasets were extremely imbalanced in their respective nature. In classification dataset, only ~2\% images were malignant and in segmentation dataset, the lesion size was very small relative to image size for majority of the images.

\begin{figure}
    \centering
    \includegraphics[scale=0.5]{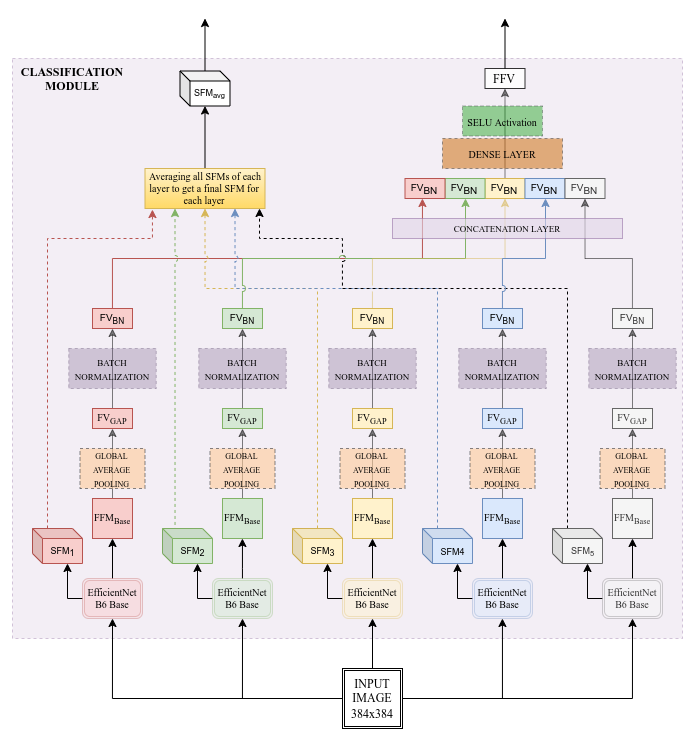}
    \caption{Ensemble Classification Module}
    \label{fig:Classification_Module}
\end{figure}

\begin{figure}
    \centering
    \includegraphics[scale=0.5]{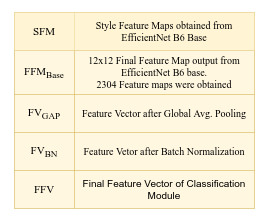}
    \caption{Key Table for Classification Model}
    \label{fig:Classification_Module_Table}
\end{figure}

\subsection{Classification Module}
Classification Module (See Figure \ref{fig:Classification_Module} for Architecture and Figure \ref{fig:Classification_Module_Table} for Keys) outputs the final feature vector and the final gram matrices of input image which is later used by RIS module to obtain similarity scores between the input image and images present in the database. Classification Module is an ensemble of classification models. Ensembling helped the model in learning more general features. The search for the best classification model was limited to \textbf{EfficientNet Family} by \citet{tan2020efficientnet} because of their SOTA performance. Out of all 7 EfficientNet Architectures from B0-B7, EfficientNet-B6 performed the best by achieving AUC of \textasciitilde 0.90, hence, it was chosen as the backbone for the ensemble classification model. In the ensemble classification model, 5 EfficientNet-B6 models were used. The classification ensemble outputs 7 gram matrices and a feature vector of length 1024. For getting gram matrices to calculate style loss, final addition layer of each of the 7 blocks present in EfficientNet-B6 was chosen. The obtained set of feature maps for each layer was converted to a gram matrix, thus giving a set of 7 gram matrices from each of the 5 EfficientNet-B6 models. The gram matrices of each layer from the 5 models were averaged out to get a final gram matrix for each layer, thus giving a final set of 7 gram matrices as output.

Due to extreme imbalance present in the classification dataset, \textbf{Focal Loss} by \citet{lin2018focal} was used for training the individual EfficientNet-B6 models. For the training the ensemble, the EfficientNet-B6 models were frozen and a top dense ML head having sigmoid as it's activation function was used. Each EfficientNet-B6 model outputs 2304 12x12 feature maps, which are then passed through Global Average Pooling Layer, Batch Normalization Layer and finally outputs of all models are concatenated. This Concatenated Feature Vector is converted into a final feature vector of length 1024 by passing it through a Dense Layer, which uses SELU as its activation function. The final feature vector is passed to the ML Head. The complete ensemble model achieved an AUC of \textasciitilde94 on the test set which is considerably better than a single EfficientNet-B6 model. After the ensemble model was trained, the top dense layer or ML head was removed to get a feature extraction model. This feature extraction model was used as the classification module which outputs the final feature vector and final set of gram matrices that gets passed to the RIS module.

\begin{figure}
    \centering
    \includegraphics[scale=0.4]{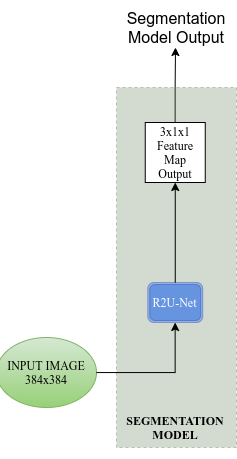}
    \caption{Segmentation Module}
    \label{fig:Segmentation_Module}
\end{figure}

\begin{figure}
    \centering
    \includegraphics[scale=0.4]{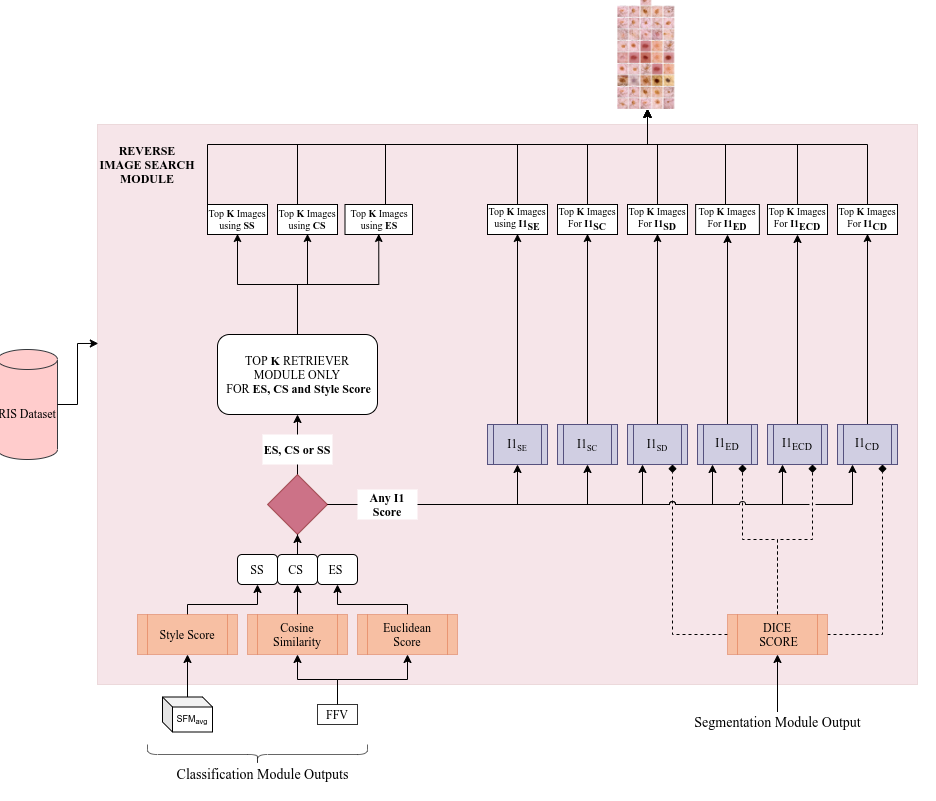}
    \caption{Reverse Image Search Module}
    \label{fig:RIS_Module}
\end{figure}

\subsection{Segmentation Module}
Segmentation Module (See Figure \ref{fig:Segmentation_Module}) outputs the semantic segmented output of the input image, later gets used by RIS module to calculate Dice Score between input image and images present in the dataset. Like classification dataset, segmentation dataset was also imbalanced, the lesion size in majority of the cases was extremely small relative to the image, hence, \textbf{Tversky Focal loss} by \citet{abraham2018novel} was used for training the segmentation models. 3 segmentation architectures were tried out: 1)\textbf{U-Net}\cite{ronneberger2015unet}, 2)\textbf{RU-Net}\cite{alom2018recurrent} and 3) \textbf{R2U-Net}\cite{alom2018recurrent}. Out of the 3 architectures, R2U-Net performed the best by achieving Dice score of \textasciitilde0.98, hence was selected as the segmentation module.

\subsection{Reverse Image Search Module}
The Reverse Image Search (RIS) Module (See Figure \ref{fig:RIS_Module}) handles the extraction of top k similar images for all the similarity measures. Classification dataset (prior to augmentation) had XX images, so, in-order to reduce inference time, the feature vectors, gram matrices and segmented output of each image was obtained beforehand using the research system and stored alongside the images in a separate database called RIS Dataset. So, RIS Dataset consisted of - images, feature vectors, gram matrices and semantic segmented output of each image.\\
The RIS module takes the feature vector, gram matrices and segmented image output from classification module and segmentation module respectively. The inputs are passed to their respective similarity measure modules which output their respective similarity scores. This scores were then used to fetch the top k similar images from RIS dataset. 

\section{Results}
\subsection{Evaluation Strategy}
For comparing the measures, 50 randomly selected images from test set of classification dataset were used. The output of each test image underwent a comprehensive manual evaluation. For this research, k=5 was chosen, so, top 5 similar images were extracted for a particular test image. Then, each output image will contribute 100/5\% = 20\% towards the accuracy of predicting similar images for that particular test image. eg: For an Image A, out of top 5 images, if only 3 are determined to be similar to image A, then the accuracy of RIS system for image A is 60\%. Using this evaluation strategy, accuracy for all 50 test images was calculated for each of the 5 similarity measure and average accuracy over 50 test images for each similarity was reported.

Sometimes, it was easy to tell whether an output image is similar to input image. But, when there was difficulty in classifying an output image as similar, careful inspection was performed using the below given criterias.
\begin{itemize}
    \item Color of the lesions
    \item Spread Density of the lesions
    \item Texture of the lesions
    \item Shape of the lesions
    \item Size of the lesions
    \item Capturing of any unique characteristics like hair etc.
    \item Color of the background
    \item Location of lesions in the images
\end{itemize}

Criterias were given different weightage. Color, spread density and texture were given the highest priority, whereas location of lesion in the image was given the least priority. The reason for giving different weightage to criterias is because it can be that lesion present in two images can have-
\begin{itemize}[align=left]
    \item[\textbf{Different Shape}] - There is no exact shape in which melanoma will spread. Image having same color, spread and texture but different shape as the input image will be considered similar.
    \item[\textbf{Different Size}] - It can be that the input image was taken when melanoma had just begun to spread whereas images present in the database were taken after considerable amount of time had pass or vice-versa. In this case, the color, spread and texture of lesions in two images can be nearly same, but size will change overtime as melanoma spreads.
    \item[\textbf{Different Unique Characteristics}] - It can be that the input image has hair present and the images in database does not have any hair present, but rest of the features are nearly same in both the images. 
    \item[\textbf{Different Location of lesion in the image}] - It can be that two lesions in input and output images respectively have the same color, spread density, texture, shape and size, but one is at top-right corner and other is at bottom-left corner. Hence, location of lesion in the image was given the least priority.
\end{itemize}

Hence, if the color, spread density and texture of the lesions present in input and output images were found to be similar, then the images are considered similar. Otherwise, rest of the criterias were also looked into.

\subsection{Observation}
Euclidean Score was taken as the baseline for this research out of the two, euclidean score and cosine similarity, due to the latter being higher. The results are shown in Table-\ref{tab:Results}. It can be observed that Style Score performed extremely well having an accuracy increase of 50\% over the baseline. The Style Score used multiple feature maps from different levels to determine the similarity, thus was able to perform well. The I1 Scores which used Style Score also performed better than the baseline by a small margin. The I1 Scores where Dice co-efficient was used performed extremely poorly. It was found that, although, addition of Dice Coefficient resulted in the increase of importance being given to size, location and shape features, it also undermined color, texture, and density features which was not ideal given that these 3 features were of utmost importance. From manual evaluation, it was also observed that I1 Scores using Style Scores performed well when the input images had well defined boundaries, and the color of lesion had a high frequency in the RIS database, but otherwise the traditional approaches added noise to the prediction, so these I1 Scores were not able to surpass Style Score.

\begin{table}
 \caption{Results}
  \centering
  \begin{tabular}{lll}
    \toprule
    %\multicolumn{2}{c}{Part}                   \\
    \cmidrule(r){1-2}
    Similarity Measure     & Success Rate                \\
    \midrule
    Euclidean Score        &    40.80\%                  \\
    Cosine Similarity      &    34.00\%                  \\
    Style Score            &    \textbf{61.20\%}         \\ 
    I1\textsubscript{ED}   &    18.00\%                  \\
    I1\textsubscript{CD}   &    10.00\%                  \\
    I1\textsubscript{ECD}  &    18.40\%                  \\
    I1\textsubscript{SE}   &    44.40\%                  \\
    I1\textsubscript{SC}   &    46.40\%                  \\
    I1\textsubscript{SD}   &    06.00\%                  \\
    \midrule
    Style Score            &    \textbf{61.20\%}         \\
    \bottomrule
  \end{tabular}
  \label{tab:Results}
\end{table}

\section{Conclusion}
The paper explores a few CBIR approaches for melanoma based on style transfer and semantic image segmentation using a novel similarity measure called I1-Score. It was found that style transfer approach which used Style Loss generalised well and performed better than the baseline measure - Euclidean Distance. But, the Semantic Image Segmentation based approach which used dice coefficient didn't fair well and showed huge accuracy drop. Although this approach resulted in increased weightage of size, shape and location, it also decreased the weightage of color, density and texture features which are even more important. Hence, this research shows that creating a CBIR systerm for melanoma using Style Loss as a similarity measure can outperform traditional approaches but adding Dice-Coefficient via I1-Score will give extremely inaccurate output.

\bibliographystyle{abbrvnat}
\bibliography{Melanoma_RIS_paper}  

\end{document}